\documentclass{IEEEcsmag}

\usepackage[colorlinks,urlcolor=blue,linkcolor=blue,citecolor=blue]{hyperref}

\usepackage{upmath}
\usepackage{graphicx}
\usepackage{epsfig, epsf, subfigure}
\usepackage[superscript]{cite}
\usepackage{amsmath,amssymb,amsfonts}
\usepackage{algorithmic}
\usepackage{textcomp}
\usepackage{amsopn}
\usepackage{etoolbox}
\usepackage{color}
\usepackage{threeparttable}
\usepackage{dsfont}
\usepackage{comment}
\usepackage{algorithm,algorithmic}
\usepackage{caption}
\usepackage{makecell}
\usepackage{booktabs}
\DeclareGraphicsExtensions {.pdf,.png,.jpg}

\jvol{XX}
\jnum{XX}
\paper{8}
\jmonth{May/June}
\jname{IT Professional}
\pubyear{2019}

\begin{document}

\sptitle{Department: Head}
\editor{Editor: Name, xxxx@email}

\title{\fontsize{27}{28}\selectfont Proxy Experience Replay: Federated Distillation for Distributed Reinforcement Learning}

\author{%
	Han Cha\IEEEauthorrefmark{1},
	Jihong Park\IEEEauthorrefmark{2},
	Hyesung Kim\IEEEauthorrefmark{3}, \\
	Mehdi Bennis\IEEEauthorrefmark{4}, and
	Seong-Lyun Kim\IEEEauthorrefmark{1}\IEEEauthorrefmark{5}
	\\
	\IEEEauthorblockA{%
		\IEEEauthorrefmark{1}Yonsei University
	}
	\IEEEauthorblockA{%
		\IEEEauthorrefmark{2}Deakin University
	}
	\IEEEauthorblockA{%
		\IEEEauthorrefmark{3}Samsung Electronics Co., Ltd.
	}
	\IEEEauthorblockA{%
		\IEEEauthorrefmark{4}University of Oulu
	}
	\IEEEauthorblockA{%
		\IEEEauthorrefmark{5}\small{Corresponding author. Contact him at slkim@yonsei.ac.kr.}
	}
}

%
%
%
%

\markboth{Department Head}{Paper title}

\begin{abstract}
Traditional distributed deep reinforcement learning (RL) commonly relies on exchanging the experience replay memory (RM) of each agent. Since the RM contains all state observations and action policy history, it may incur huge communication overhead while violating the privacy of each agent. Alternatively, this article presents a communication-efficient and privacy-preserving distributed RL framework, coined \emph{federated reinforcement distillation (FRD)}. In FRD, each agent exchanges its proxy experience replay memory (ProxRM), in which policies are locally averaged with respect to proxy states clustering actual states. To provide FRD design insights, we present ablation studies on the impact of ProxRM structures, neural network architectures, and communication intervals. Furthermore, we propose an improved version of FRD, coined \emph{mixup augmented FRD (MixFRD)}, in which ProxRM is interpolated using the mixup data augmentation algorithm. Simulations in a Cartpole environment validate the effectiveness of MixFRD in reducing the variance of mission completion time and communication cost, compared to the benchmark schemes, vanilla FRD, federated reinforcement learning (FRL), and policy distillation (PD).

\end{abstract}

\maketitle

\begin{figure*}
	\center
	\subfigure[Policy distillation.] {\includegraphics[width=11cm]{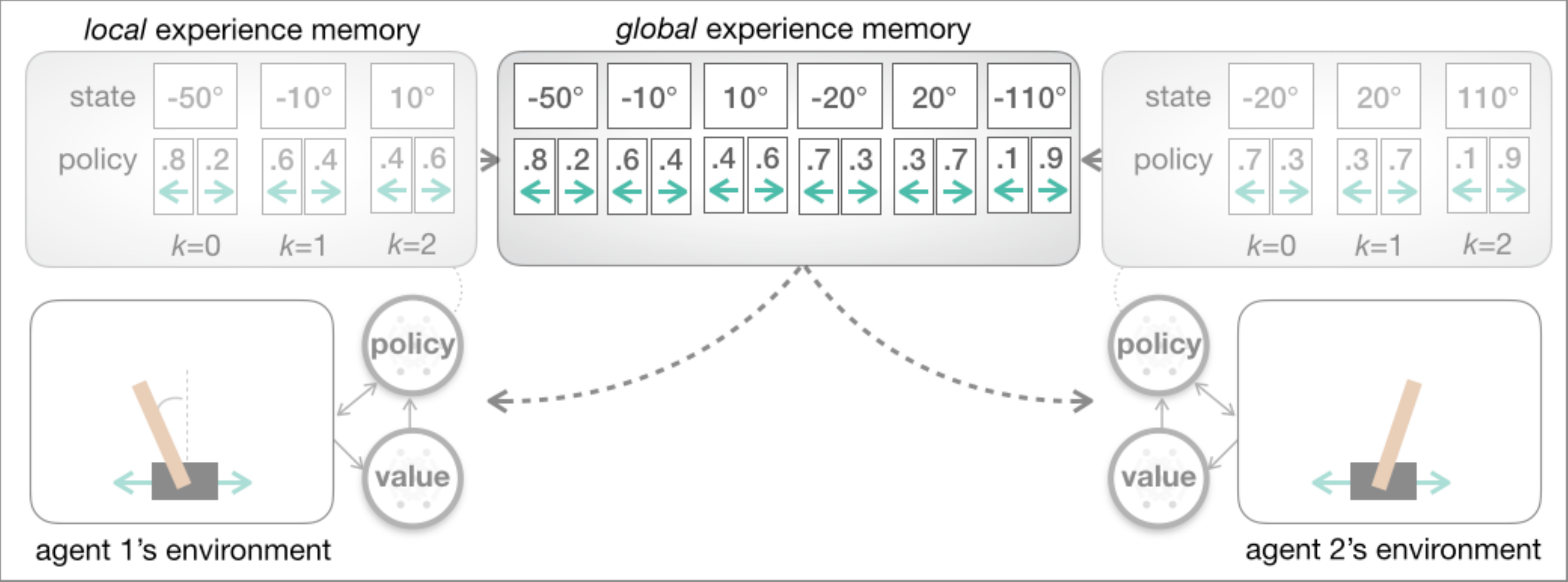}}
	\subfigure[Federated reinforcement distillation with proxy replay memory.] {\includegraphics[width=11cm]{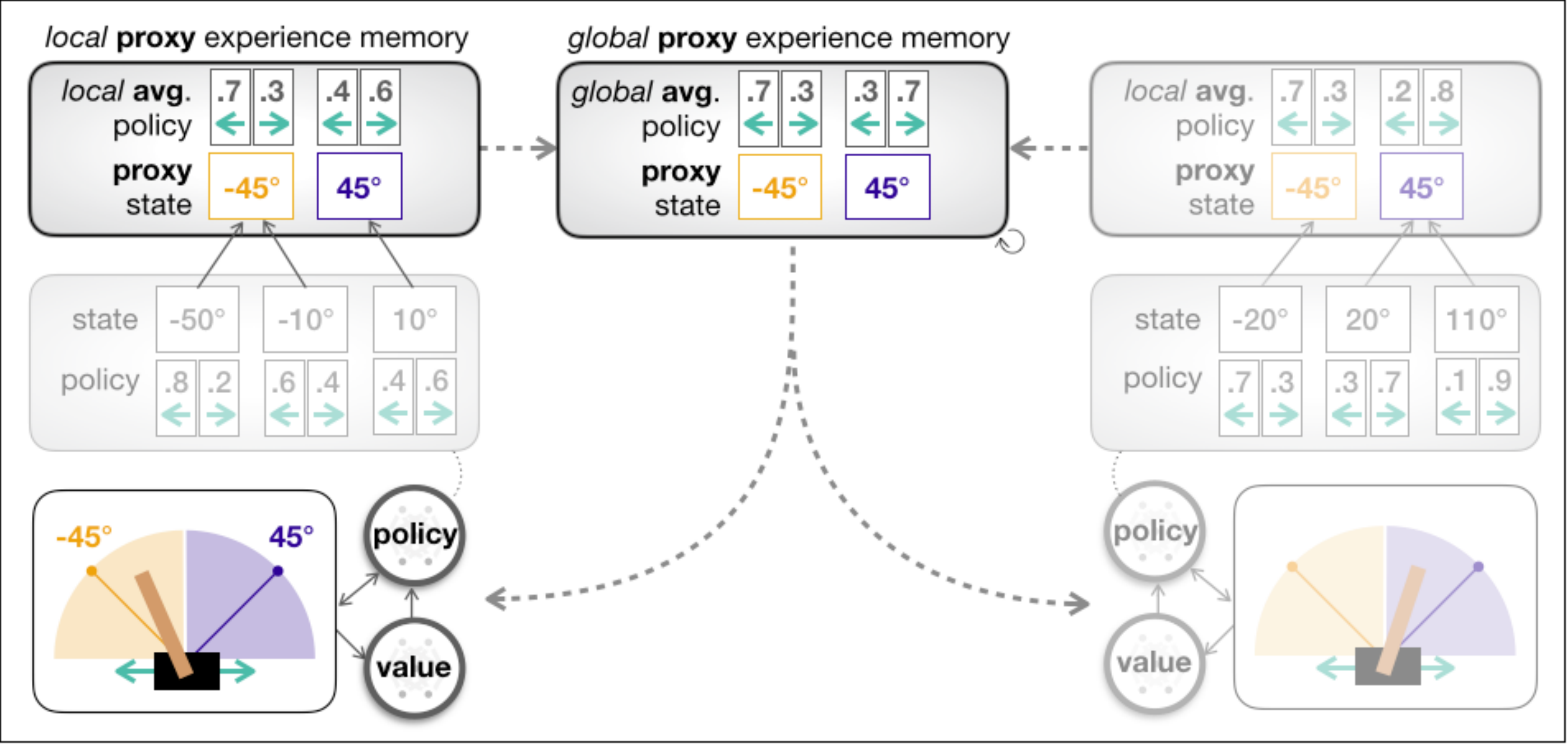}}
	\caption{A comprehensive structure of distributed deep RL frameworks.} \label{Fig:structure}
\end{figure*}

\chapterinitial{Learning} from collective experience is a hallmark of intelligent agents powered by distributed deep reinforcement learning (RL), ranging from autonomous drones\cite{Park19} to self-controlled robots in smart factories.\cite{Hamid19} Policy distillation (PD) is one of the most popular distributed RL methods.\cite{Rusu16} As shown in Fig. \ref{Fig:structure}(a), in PD, each agent has a neural network (NN) model and its \emph{local replay memory} storing the action policies taken at the agent's observed states. This local replay memory is exchanged across agents. By replaying the globally collected replay memory, every agent trains its NN model while reflecting the experiences of other agents. 

However, in practice, local experience memories can be privacy-sensitive, which prohibits their exchanges. As an example, for autonomous surveillance drones operated by multiple operators, these drones are willing to train their NNs collectively, but without revealing their private observations and policy records. Furthermore, the dimension of possible states can be large, as in high-precision robot manipulation. This leads to huge replay memory sizes, and exchanging them is ill-suited for agents' operations in real-time.

Alternatively, our prior work proposed a privacy-preserving and communication-efficient distributed RL framework, coined \emph{federated reinforcement distillation (FRD)}.\cite{Cha19} As illustrated by Fig. \ref{Fig:structure}(b), in FRD, each agent exchanges its \emph{proxy experience memory replay memory (ProxRM)} consisting of locally averaged policies with respect to proxy states. Under this memory structure, policies are locally averaged over time, while actual states are clustered into the nearest proxy states. Exchanging proxy replay memory can thereby preserve privacy while reducing the communication payload size.

\begin{figure*}[h!]
	\centering
	\subfigure[Setting  1.] {\includegraphics[width=2.9cm]{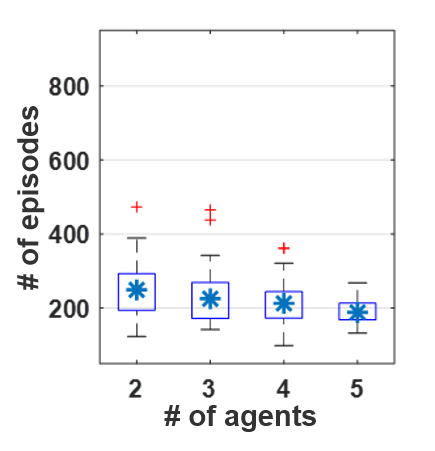}}
	\subfigure[Setting  2.] {\includegraphics[width=2.9cm]{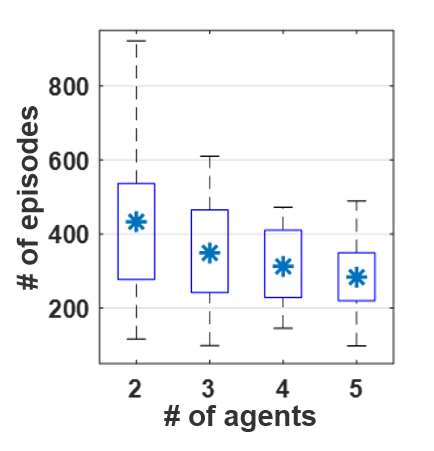}}
	\subfigure[Setting  3.] {\includegraphics[width=2.9cm]{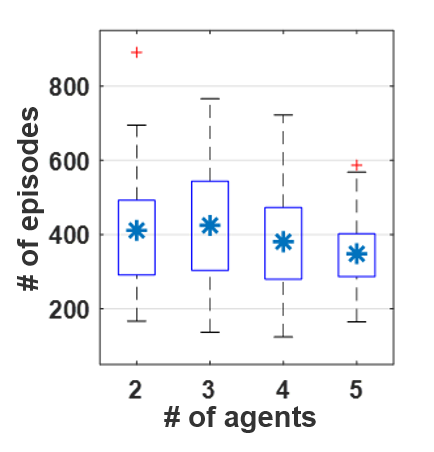}}
	\subfigure[Setting  4.] {\includegraphics[width=2.9cm]{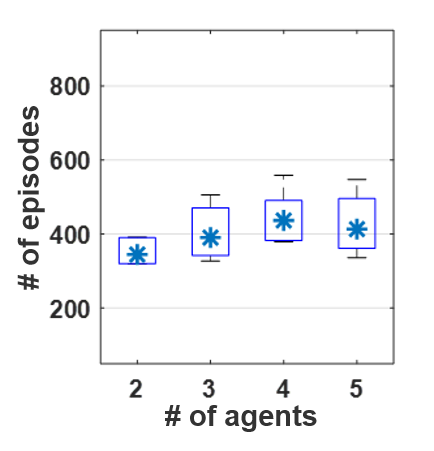}}
	\subfigure[Setting  5.] {\includegraphics[width=2.9cm]{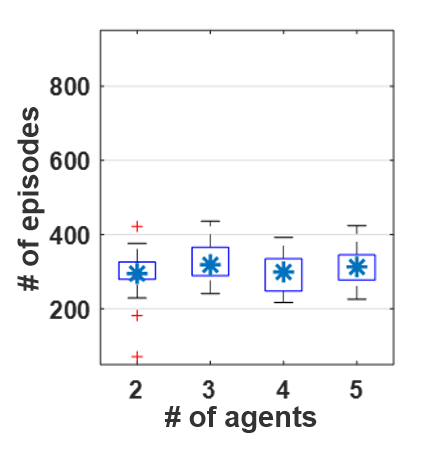}}
	\setlength{\belowcaptionskip}{-10pt}
	\setlength{\abovecaptionskip}{-2pt}
	\caption{Ablation study on FRD under different hyperparameter settings provided in Table~\ref{Tab:Parameters}. } \label{Fig:extsim}
\end{figure*}

In this article, we aim to provide an FRD design guideline, shedding light on how coarse the proxy states are and how frequent the proxy experience memories are exchanged under which NN architectures. Furthermore, we propose an improved version of FRD, named MixFRD, in which the globally collected proxy replay memory is interpolated using the mixup data augmentation algorithm.\cite{Zhang18} Finally, we demonstrate the effectiveness of MixFRD by comparing it with two baseline algorithms, PD and federated reinforcement learning (FRL),\cite{McMahan17} in terms of mission completion time and communication payload size.

\section{Federated Reinforcement Distillation (FRD)}
In FRD, for a given environment and mission, each agent locally updates its NN model by observing the states while taking actions, and stores the average action policies per proxy state to construct a ProxRM that is periodically exchanged across agents. For the sake of explanation, hereafter, we focus on the Cartpole environment and describe FRD operations as detailed next.

\subsection{Environmental Settings}
We evaluate the performance of FRD using the \emph{Cartpole-v1} in the OpenAI gym environment.\cite{Brockman16} The objective of this game is to make the pole attached to a cart upright as long as possible, by moving the cart left or right. The RL agent gets a score of $+1$ for every time slot that the pole remains upright during an episode. Every episode ends if (i) the pole fall down; (ii) the cart moves outside a pre-defined range; or (iii) the score reaches 500. Playing the game with multiple episodes, agent completes their mission when any of them first reaches an average score of 490, where the average is taken across 10 latest episodes. We define a mission completion time as the number of played episodes when the mission is accomplished.

\subsection{Standalone and Distributed RL Operations}

Each agent under study runs the advantage actor-critic (A2C) RL algorithm\cite{Mnih16}. In A2C, there is a pair of NNs, in which a policy NN determines the agent's optimal action policies of input states, and the value NN examines the optimality of the policies. To this end, for each action decision, the value NN produces  value that is used for calculating an advantage. When the advantage is near zero, the determined action is closer to the optimal action compared to other possible actions. Therefore, the policy NN is trained by minimizing the advantage produced by the value NN. While running the aforementioned standalone RL operations, agents periodically exchange their policy NN outputs and inputs through ProxRMs, whose structures are elaborated in the next subsection. Each agent replays the ProxRM similar to the training procedure of supervised learning.




\subsection{Proxy Experience Replay Memory Structures}
In the Cartpole environment, each state is described by four components: cart location, cart velocity, pole angle, and the angular velocity of the pole. To construct a ProxRM, the state is evenly divided into a number $S$ of clustered values per each component and give an index number to each cluster. Along with the state clusters, we categorize the RM following the nearest distance rule and average out the policies. The proxy states are defined by the middle value of each state cluster. For example, the proxy state is -45$^\circ$ if the state cluster is [-90$^\circ$, 0$^\circ$), as illustrated in Fig. \ref{Fig:structure}(b). Finally, we have ProxRM with (proxy state, averaged policies) tuples.



\begin{table}
	\centering
	\caption{Hyperparameter settings of FRD.}
	\label{Tab:Parameters}
	\resizebox{
		0.48\textwidth}{!}{
		\begin{tabular}{c|c|c|c|c}
			\toprule
			\thead{Setting } & \thead{ \# of sections for \\ state components ($S$)} & \thead{communication \\ period ($ E $)} & \thead{ \# of weights \\ per layer ($ n $)} & \thead{ \# of hidden \\ layers ($\ell$)}\\
			\midrule
			\textit{1} & $100$ & 25 &  \textbf{24} & 2 \\
			\textit{2} & $\mathbf{100}$ & 25 & \textbf{100} & 2 \\
			\textit{3} & {$\mathbf{50}$} & 25  & 100 & 2 \\
			\textit{4} & $100$ & \textbf{10}  & 24 & 1 \\
			\textit{5} & $100$ & \textbf{50}  & 24 & 1 \\
			\bottomrule          
		\end{tabular}
	} 
\end{table}

\section{Ablation Study on FRD}
To provide FRD design insights, in this section, we study the impact of ProxRM structures, NN architectures, and communication protocols. Specifically, we consider five parameters: the number $S$ of clusters per state component, ProxRM exchanging period $E$, the number $L$ of a multi-layer perceptron (MLP) NN hidden layers, and the number $n$ of weights per layer. 

Under the settings described in Table~\ref{Tab:Parameters}, Fig.~\ref{Fig:extsim} shows the following impacts of FRD design parameters.

\begin{itemize}
	\item Too small number $S$ of clusters makes FRD fail to obtain federation gains. As opposed to Fig.~\ref{Fig:extsim}(b), the mission completion time in Fig.~\ref{Fig:extsim}(d) does not decrease with the number of agents due to too small $S$. Since $S$ determines the communication payload sizes in FRD, there is a \emph{trade-off between communication efficiency and scalability}.
	
	\item Too frequent ProxRM exchanges may be harmful, as observed by Fig.~\ref{Fig:extsim}(d) compared to (e). Without a sufficient number of local RL iterations, the observed states are hardly correlated across agents, negating the effectiveness of their federation. Since $E$ determines the number of communication rounds, it is important to \emph{balance the local RL iterations and ProxRM communication interval}.
	
	\item A larger NN often has higher sample complexity, and does not always outperform a smaller NN, as shown in Fig.~\ref{Fig:extsim}(b) compared to (a). Nonetheless, for a larger NN, federation gain is higher, since exchanged ProxRMs more overcomes the lack of training samples. It is therefore crucial to \emph{optimize the number of federating agents subject to their NN architectures.}

	

	
\end{itemize}

\section{Mixup Augmented FRD (MixFRD)}
In this section, we propose an enhanced version of FRD coined \textit{MixFRD}, which utilizes the augmentation scheme to enrich ProxRM. With few agents, the exploration of the environment may be insufficient compared to that of many agents. The lack of exploration of ProxRM causes performance instability in the FRD framework. We handle this problem by applying the data augmentation algorithm by generating synthetic ProxRM with downloaded global ProxRM at the local agent. 

A \emph{mixup}\cite{Zhang18} algorithm is one of the most promising augmentation schemes. The main purpose of mixup is to enhance the generalization capability of NN by generating unseen data samples. The mixup algorithm creates a new data sample and its corresponding label by a linear combination of the randomly picked two existing data samples. In general, but not limited, the portion of two data samples follows the Beta distribution with coefficient $0<\alpha=\beta<1$. The portion is selected near 0 or 1 value in this setting.

To apply mixup to the FRD framework, there are two considerations about selecting what kind of two samples in ProxRM and the portion of linear combination. At first, if we pick two memories randomly, they may be uncorrelated. As a result, augmented proxy state and averaged action policy are not valid for experience memory. This augmented sample causes severe performance degradation if it used for training agent model. Second, if we pick the portion of linear combination as same as the general setting of mixup, the augmented proxy state and averaged action policy is similar to either original samples, which is no help for expanding the ProxRM. Therefore, we need to select highly correlated ProxRM samples and the portion reflecting the samples equally. 

\begin{figure} [t]
	\center
	\includegraphics[width=7.6cm]{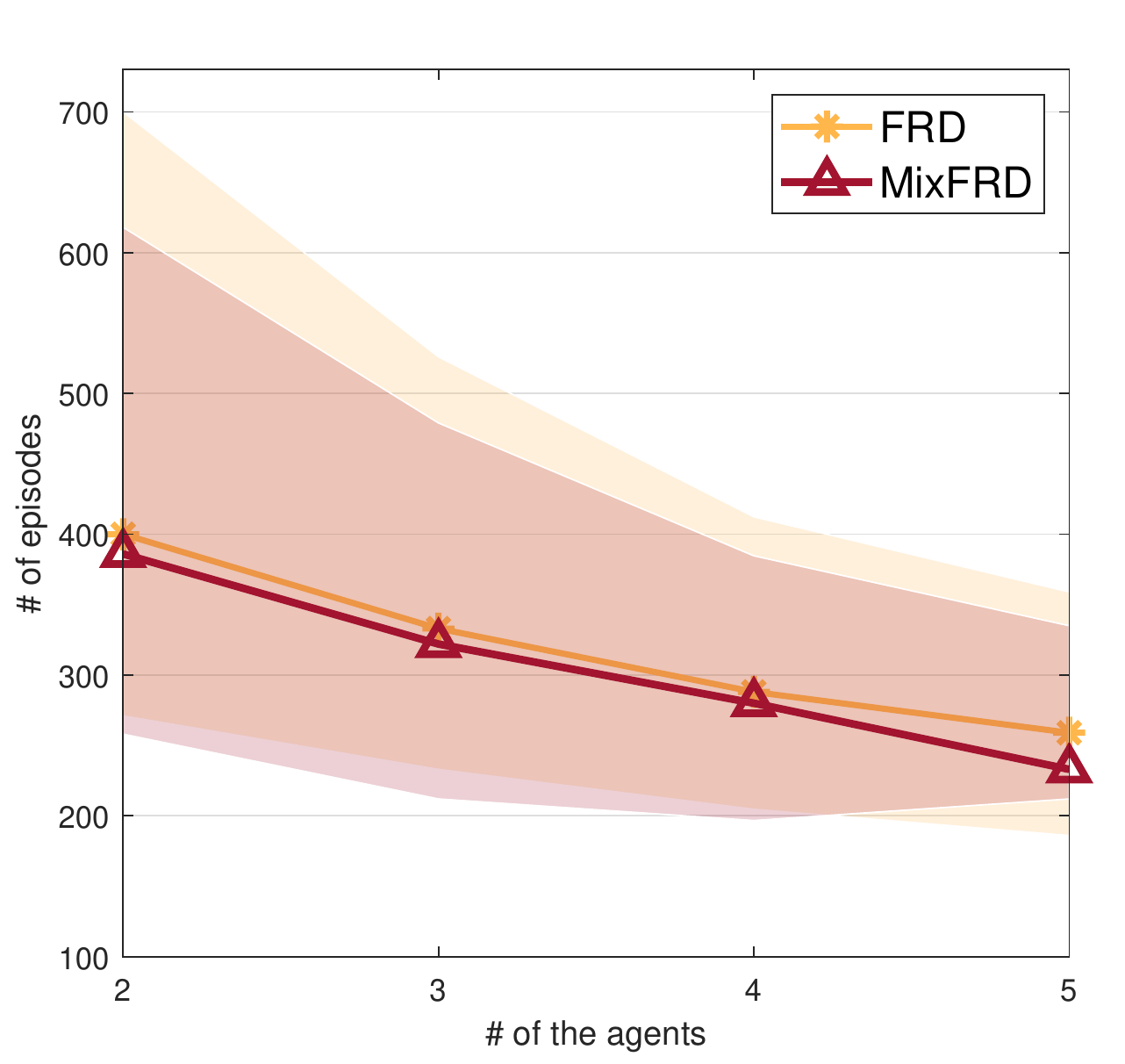}
	\caption{Mission completion time comparison between FRD and MixFRD ($S = 30$, $E = 25$, $n = 50$, $l = 2$).} \label{Fig:FRD_MixFRD}
\end{figure}

In the Cartpole environment, the angle of the pole is the most relevant information to accomplish the group mission among the four components. To ensure the correlation between two memories, we sort the ProxRM along with the angle of the pole and select two adjacent replay memories. With this memory, we conduct the mixup with a portion of $0.5$. We can say that the newly constructed ProxRM is an interpolated version of the original one. In Fig. \ref{Fig:FRD_MixFRD}, the MixFRD shows lower variance of mission completion time compared to that of FRD, as expected. In general, the correlation among the ProxRM can be calculated by measuring the distance between each state clusters, e.g., Euclidean distance, cosine distance, Jaccard distance, etc.

\begin{figure} [t]
	\center
	\includegraphics[width=7.6cm]{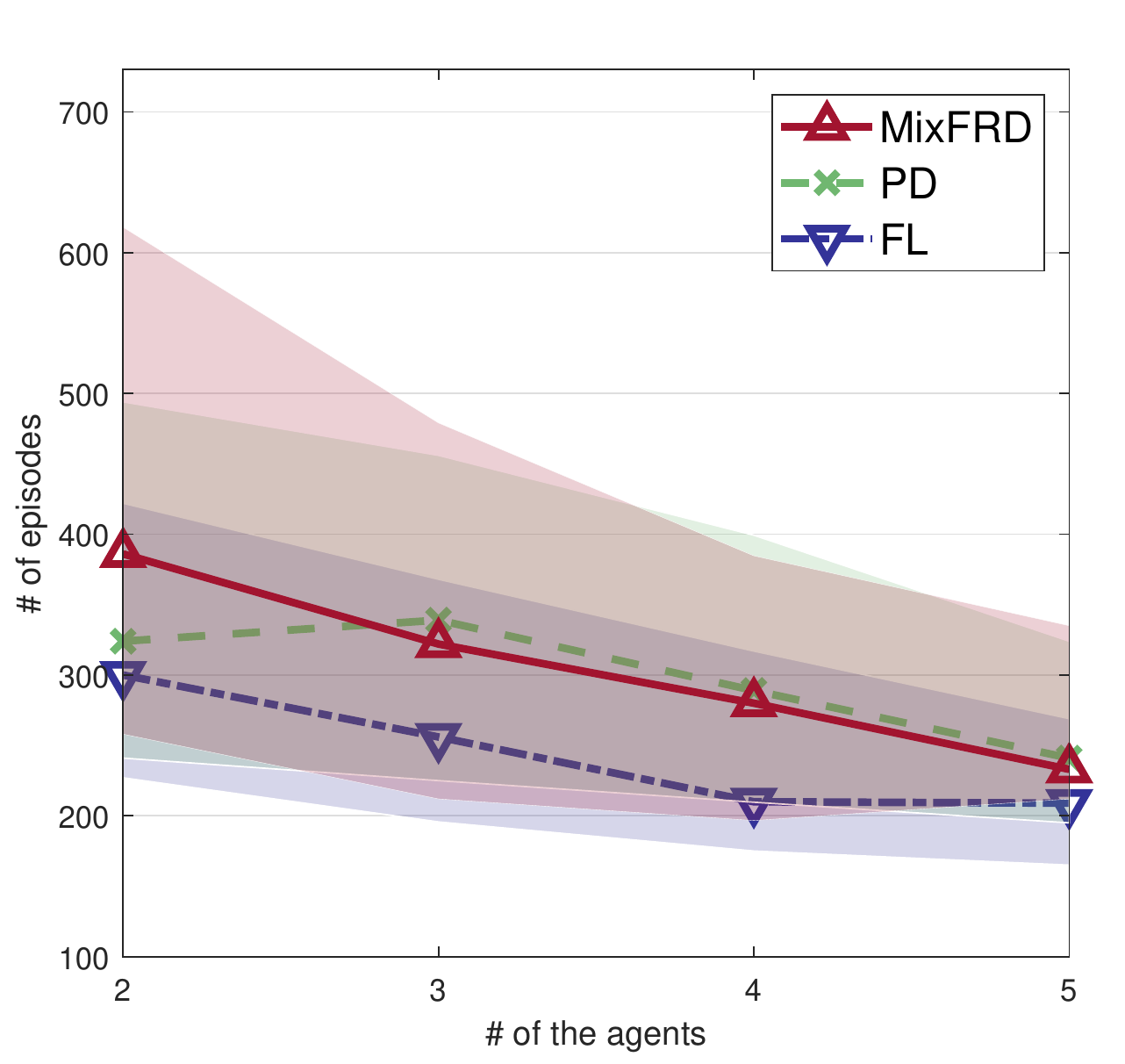}
	\caption{Mission completion time comparisons among MixFRD, PD, and FRL.} \label{Fig:3comp}
\end{figure}
	
\section{Performance Evaluation}
We numerically compare the performance of the proposed MixFRD framework with two baseline distributed deep RL frameworks: policy distillation (PD)\cite{Rusu16} and federated reinforcement learning (FRL).\cite{McMahan17} Each framework uses a different form of knowledge: MixFRD exchanges ProxRM, PD exchanges RM, and FRL exchanges weight parameters of local NN. The sizes of each knowledge are denoted by $M^p$, $M$, $W(n,l)$, respectively. The subscripts $L, G$ denote the local and global knowledge, respectively. Note that agents using FRL reflect global knowledge by exchanging the local weights to the global weight parameters.

We evaluate the performance with two metrics: (i) mission completion time, (ii) payload size per knowledge exchange. We conduct the simulation under the following settings: $S = 30$, $E = 25$, $n = 50$, $l = 2$. The lines represent the median, and the colored area is between the top-25 percentile and the top-75 percentile of each case. In the latter evaluation, we conduct the simulation for two agents and adjust $n$ of policy network only.

The ProxRM size $M^p$ is determined by the number of distinct state observations, and upper bounded by the entire cluster dimension $S^4$. As $S$ goes to infinity, ProxRM converges to RM, i.e., MixFRD is identical with PD. Note that ProxRM size $M^p$ increases with $S$ only if the preceding state observations are sufficiently diverse. 

\begin{figure} [t]
	\center
	\includegraphics[width=7.6cm]{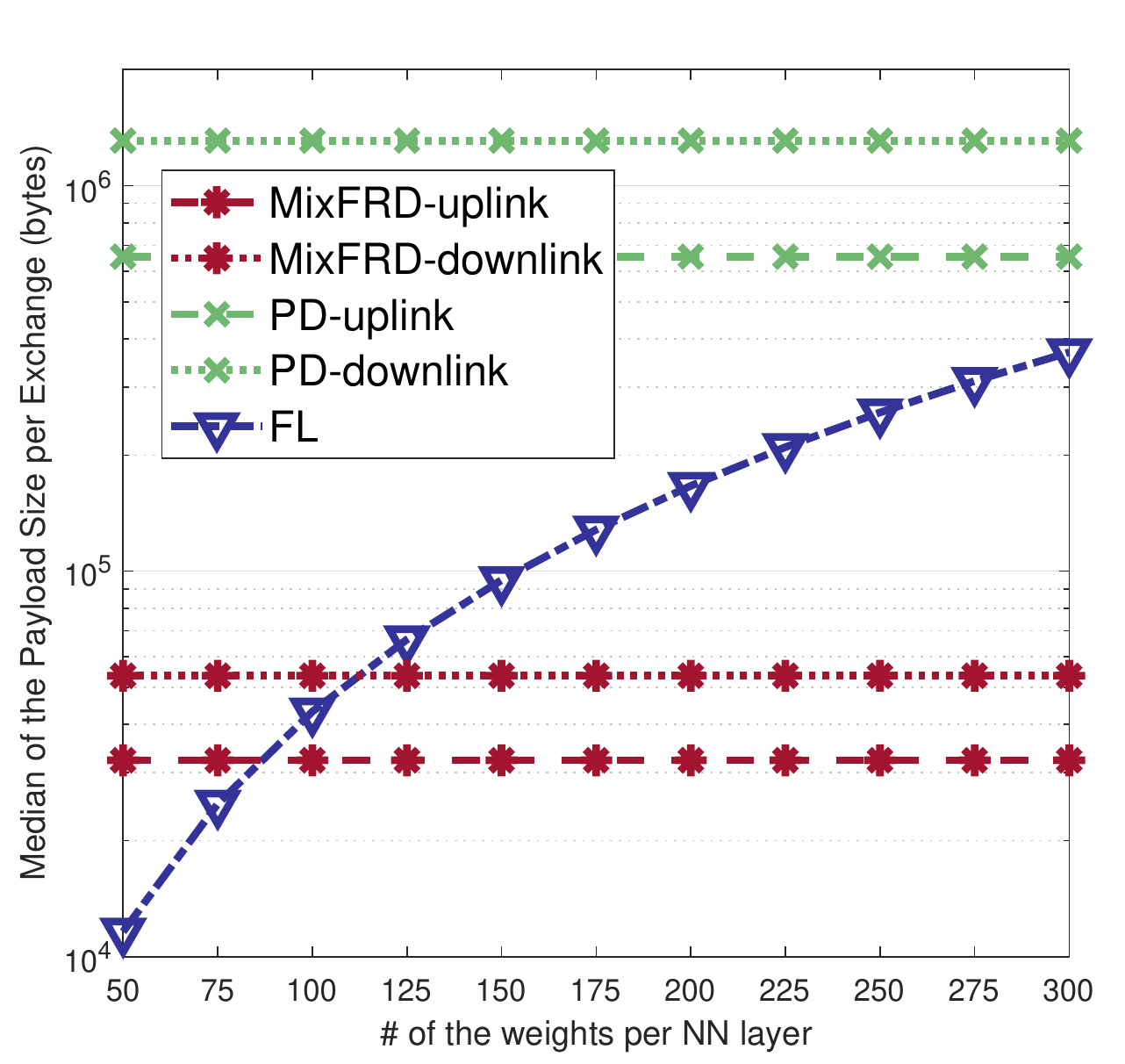}
	\caption{Uplink and downlink payload size comparison for two agents with the server.} \label{Fig:payload}
\end{figure}

\begin{table}
	\centering
	\caption{Communication payload sizes.}
	\renewcommand{\arraystretch}{1.2}
	\resizebox{
		0.4\textwidth}{!}{
		\begin{tabular}{c|c}
			\toprule
			\thead{ Cases } & \thead{ Payload size (byte)} \\ 
			\midrule
			MixFRD-uplink & $b_{p}\times M_L^{p}$ \\ 
			MixFRD-downlink & $b_{p}\times M_G^{p}$ \\
			PD-uplink & $b_{rm}\times M_L$ \\
			PD-downlink & $b_{rm}\times M_G$ \\
			FRL & $b_{w}\times W(n,l)$ \\
			\bottomrule          
		\end{tabular} \label{Tab:payload}
	} 
\end{table}

The communication payload sizes of each framework are defined in Table \ref{Tab:payload}. We define the unit size of ProxRM, RM, and weights of FRL as $b_p=12$, $b_{rm}=24$, and $b_w=4$, respectively. RM comprise of six float components, which are four state components and two action policies, respectively. ProxRM comprises of one signed integer and two float components, which are index of state cluster and two action policies, respectively. The weight of FRL is float integer, which is the weight parameter. In the case of MixFRD, the server defines the ProxRM structure in advance, in which all the agents and server have information about the entire state cluster index and corresponding proxy state. Therefore, the agents in MixFRD share only the state cluster index.

	\subsection{Comparison with PD}
	In Fig. \ref{Fig:3comp}, the proposed MixFRD has a compatible performance with the PD when the number of cooperating agents is above two. Furthermore, the MixFRD has far less payload size compared to that of the PD, as in Fig. \ref{Fig:payload}, which illustrates two-agents case. the gap between the two frameworks cannot be narrowed as the number of cooperating agents increases. The payload sizes of the MixFRD and PD have a positive correlation with the size of explored state space of learning agent. For example, when the agent holds the pole straight, the observed state space is monotonous. During that case, the learning agent gets similar RMs, which cause payload size explosion. By utilizing MixFRD, the agents average out these RMs with ProxRMs, which has far fewer number compared to the RM. Furthermore, as the cooperating agents increase, the number of global RM increases exponentially. Therefore, MixFRD can replace PD by exchanging the RM payload size.
	
	\subsection{Comparison with FRL}
	In Fig. \ref{Fig:3comp}, the performance of FRL overwhelms that of the MixFRD. However, as the size of local NN increases, the payload size of FRL increases with the squared scale of the size, as presented in Fig. \ref{Fig:payload}. The MixFRD framework has an advantage in payload size constrained scenarios, e.g., cooperating through wireless channels like swarm of drones, automated process of smart factories, etc. Furthermore, all the agents require homogeneous local NN structure to use FRL, which is a strong assumption due to the diversity of computational capability of . The MixFRD can provide compatible distributed RL framework among agents which have heterogeneous local NN structure.

\section{Conclusions}

In this article, we propose a communication-efficient and privacy-preserving distributed RL framework, FRD, in which agents exchange their ProxRMs without revealing their raw RMs. Furthermore, we propose  an improved version of FRD, coined as MixFRD, by leveraging the mixup data augmentation algorithm to interpolate ProxRMs locally. Our ablation studies show that improving the FRD performance hinges on ProxRM structures, NN architectures, and communication protocols. Co-designing these system parameters can, therefore, be an interesting topic for future research.

\section{ACKNOWLEDGMENT}
This research was partly supported by a grant to Bio-Mimetic Robot Research Center Funded by Defense Acquisition Program Administration, and by Agency for Defense Development (UD190018ID), and in part by the Academy of Finland under Grant 294128. 
%

\begin{IEEEbiography}{Han Cha} 
	is currently pursuing a Ph.D. degree at the School of Electrical and Electronic Engineering, Yonsei University, Seoul, Korea. His current research includes resource management, interference-limited networks, radio hardware implementation, AI-assisted wireless communication, IoT communications, dynamic spectrum sharing, cognitive radio. He was a Visiting Researcher with the University of Oulu, in 2019. He received a B.S. degree at Yonsei University in 2015. He was awarded the IEEE DySPAN Best Demo Paper in 2018, titled "Opportunistic map based flexible hybrid duplex systems in dynamic spectrum access," as a coauthor. Contact him at chan@ramo.yonsei.ac.kr.
\end{IEEEbiography}

\begin{IEEEbiography}{Jihong Park}
	is currently a Lecturer at the School of Information Technology, Deakin University, Waurn Ponds, VIC, Australia. His recent research focus includes communication-efficient distributed machine learning and distributed ledger technology. He is also interested in ultra-reliable, ultra-dense, and mmWave system designs in 5G and beyond. He was a Post-Doctoral Researcher with Aalborg University, Denmark, from 2016 to 2017; the University of Oulu, Finland, from 2018 to 2019. He was a Visiting Researcher with the Hong Kong Polytechnic University, Hong Kong, in 2013; the KTH Royal Institute of Technology, Stockholm, Sweden, in 2015; Aalborg University, in 2016; and the New Jersey Institute of Technology, Newark, NJ, USA, in 2017. He received the B.S. and Ph.D. degrees from Yonsei University, Seoul, Korea, in 2009 and 2016, respectively. He received the 2014 IEEE GLOBECOM Student Travel Grant, the 2014 IEEE Seoul Section Student Paper Contest Bronze Prize, and the 6th IDIS-ETNEWS (The Electronic Times) Paper Contest Award sponsored by the Ministry of Science, ICT, and Future Planning of Korea. Contact him at jihong.park@deakin.edu.au.
\end{IEEEbiography}

\begin{IEEEbiography}{Hyesung Kim} 
	is currently a staff engineer with the Samsung Electronics Co. Ltd., Seoul, Korea. His research interests include edge computing/caching, 5G communications system, distributed machine learning, and blockchain. He was a Visiting Researcher with the University of Oulu, in 2018. He received the B.S. \& Ph.D. in Electrical \& Electronic Engineering from Yonsei University, Seoul, South Korea. He was awarded in the IEEE Seoul Section Student Paper Contest Bronze Prize in 2017, titled "Mean-field game-theoretic edge caching for ultra dense networks," as a first author. Contact him at hye1207@gmail.com.
\end{IEEEbiography}

\begin{IEEEbiography}{Mehdi Bennis}
	is currently an Associate Professor with the Centre for Wireless Communications, University of Oulu, Oulu, Finland, where he is also an Academy of Finland Research Fellow and the Head of the Intelligent Connectivity and Networks/Systems Group (ICON). He has coauthored one book and published more than 200 research articles in international conferences, journals, and book chapters. His current research interests include radio resource management, heterogeneous networks, game theory, and machine learning in 5G networks and beyond. Dr. Bennis was a recipient of several prestigious awards, including the 2015 Fred W. Ellersick Prize from the IEEE Communications Society, the 2016 Best Tutorial Prize from the IEEE Communications Society, the 2017 EURASIP Best Paper Award for the Journal on Wireless Communications and Networks, the All-University of Oulu Award for research, and the 2019 IEEE ComSoc Radio Communications Committee Early Achievement Award. He is an Editor of the IEEE TRANSACTIONS ON COMMUNICATIONS. Contact him at mehdi.bennis@oulu.fi
\end{IEEEbiography}

\begin{IEEEbiography}{Seong-Lyun Kim} 
	is a Professor of wireless networks at the School of Electrical \& Electronic Engineering, Yonsei University, Seoul, Korea, heading the Robotic \& Mobile Networks Laboratory (RAMO) and the Center for Flexible Radio (CFR+). He is co-directing H2020 EUK PriMO-5G project, and leading Smart Factory TF of 5G Forum, Korea. His research interest includes radio resource management, information theory in wireless networks, collective intelligence, and robotic networks. He received B.S. in Economics from Seoul National University, and M.S. \& Ph.D. in Operations Research with Application to Wireless Networks from Korea Advanced Institute of Science \& Technology (KAIST). He was an Assistant Professor of Radio Communication Systems at the Department of Signals, Sensors \& Systems, Royal Institute of Technology (KTH), Stockholm, Sweden. He was a Visiting Professor at the Control Engineering Group, Helsinki University of Technology (now Aalto), Finland, the KTH Center for Wireless Systems, and the Graduate School of Informatics, Kyoto University, Japan. He served as a technical committee member or a chair for various conferences, and an editorial board member of IEEE Transactions on Vehicular Technology, IEEE Communications Letters, Elsevier Control Engineering Practice, Elsevier ICT Express, and Journal of Communications and Network. He served as the leading guest editor of IEEE Wireless Communications and IEEE Network for wireless communications in networked robotics, and IEEE Journal on Selected Areas in Communications. He also consulted various companies in the area of wireless systems both in Korea and abroad. He published numerous papers, including the coauthored book (with Prof. Jens Zander), Radio Resource Management for Wireless Networks. He is the corresponding author for this article and contact him at slkim@yonsei.ac.kr.
\end{IEEEbiography}


\end{document}